\documentclass[11pt,a4paper]{article}
\usepackage[hyperref]{emnlp2020}
\usepackage{times}
\usepackage{latexsym}

\usepackage{fullpage}
\usepackage[top=2cm, bottom=4.5cm, left=2.5cm, right=2.5cm]{geometry}
\usepackage{amsmath,amsthm,amsfonts,amssymb,amscd}
\usepackage{lastpage}
\usepackage{enumerate}
\usepackage{fancyhdr}
\usepackage{mathrsfs}
\usepackage{xcolor}
\usepackage{graphicx}
\usepackage{adjustbox}
\usepackage{caption}
\usepackage{subcaption}
\usepackage{listings}
\usepackage{hyperref}
\usepackage{enumitem}
\usepackage{float}
\usepackage{fancyvrb}
\usepackage{enumitem}
\usepackage{bbold}
\usepackage[T1]{fontenc}

\usepackage{microtype}
\usepackage{graphicx}
\aclfinalcopy 

\title{MIX: a Multi-task Learning Approach to Solve Open-Domain Question Answering}

\author{Sofian Chaybouti, Achraf Saghe, Aymen Shabou \\
\textit{DataLab Groupe, Crédit Agricole S.A} \\
Montrouge, France}

\date{}

\begin{document}
\maketitle

\begin{abstract}

In this paper, we introduce MIX: a   multi-task deep learning approach to solve Open-Domain Question 
Answering. First, we design our system as a multi-stage pipeline of 3 building blocks: a BM25-based Retriever to reduce the search space, a RoBERTa-based Scorer, and an Extractor to rank retrieved paragraphs and extract relevant text spans, respectively. Eventually, we further improve
the computational efficiency of our system to deal with the scalability challenge: thanks to multi-task learning,  
we parallelize the close tasks solved by the Scorer and the Extractor. Our system is on par with state-of-the-art performances on the squad-open benchmark while being simpler conceptually.

\end{abstract}

\section{Introduction}

With huge quantities of natural language documents, search engines have been essential for the time saved on information retrieval tasks. Usually, deployed search engines achieve the task of ranking documents by relevance according to a query. \\
Recently, research has focused on the task of extracting the span of text that exactly matches the user's query through Machine 
Reading Comprehension and Question Answering. \\
Question Answering deals with extracting the span of text in a short paragraph that exactly answers a natural language question. Recent deep learning models based on heavy pretrained language models like BERT achieved better than human performances on this task \cite{devlin2019bert}. 
\\
One could apply QA models for the Open-Domain Question Answering paradigm, which aims to answer questions by taking many documents as knowledge sources. Two main issues emerge from this: first, applying 100M parameters language models to potentially millions of documents requires unreasonable GPU resources. Then, QA models allow to compare spans of text coming exclusively from a single paragraph while in the open-domain QA paradigm, one needs to compare spans of text coming from a wide range of documents. \\
Our system, as done in previous work, deals with the resources issue thanks to a Retriever module based on the BM25 algorithm that allows us to reduce the search space from millions of articles to a hundred paragraphs. The second issue is tackled by adding a deep learning-based Scorer module that re-ranks the paragraphs returned by the Retriever with more precision. Eventually, the Extractor module uses a QA deep learning model to extract the best span of text in the first paragraph returned by the Scorer. To avoid a heavy and hardly scalable pipeline consisting of two huge deep learning models, we parallelize the re-ranking and span extraction tasks thanks to multitask learning while maintaining high performances, which significantly reduces both memory requirements and inference time. Our system achieves state-of-the-art results on the open-squad benchmark.

\begin{figure*}[ht!]
    \includegraphics[width = \textwidth]{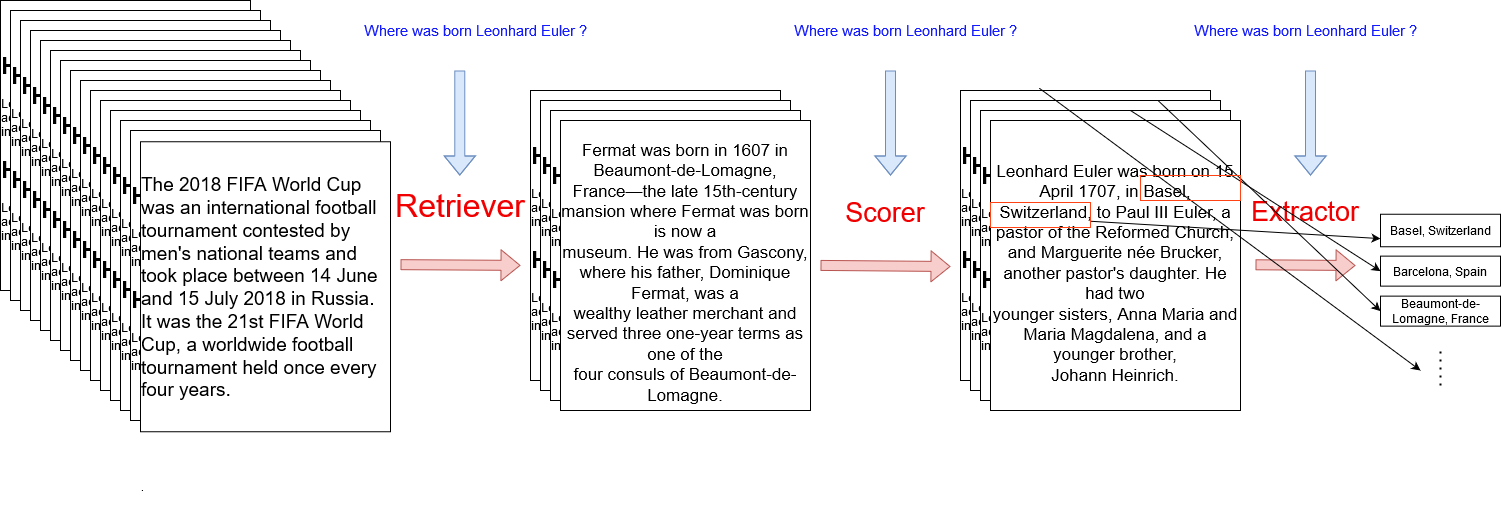}
\caption{The proposed pipeline for the open-domain QA task resolution. The search space is first reduced in this preliminary pipeline thanks to the \textit{Retriever}. Secondly, the ranking of paragraphs retrieved is refined thanks to the \textit{Scorer}. Eventually, the \textit{Extractor} allows highlighting the most relevant text spans in each paragraph.}
\label{pipeline}
\end{figure*}

\section{Background and Previous Work}

\subsection{Machine Reading Comprehension}

The construction of vast Question Answering datasets, particularly the SQuAD benchmark \cite{rajpurkar2016squad}, has led to end-to-end deep learning models successfully solving this task, for instance, \cite{seo2018bidirectional} is one of the first end-to-end model achieving impressive performances. More recently, the finetuning of powerful language models like \textit{BERT} \cite{devlin2019bert} has allowed us to achieve better than human performances on this benchmark. Some researchers have adapted the pretraining task of language models to be better adapted to the extractive question answering down-stream task like \textit{SpanBERT} \cite{joshi2020spanbert}. Eventually, all these models rely on the same paradigm: building query-aware vector representations of the words in the context. 

\subsection{Open-Domain Question Answering}

\begin{figure}[ht]
    \includegraphics[width = 0.45\textwidth]{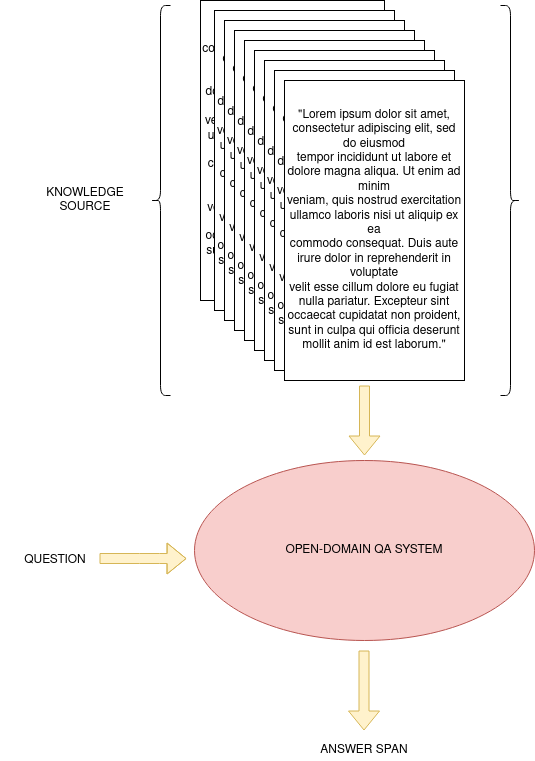}
\caption{The open-domain QA setting.}
\label{opendomainqa}
\end{figure}

\cite{chen2017reading} introduce the Open-Domain Question Answering setting (figure \ref{opendomainqa}) that aims to use the entire English Wikipedia as a knowledge source to answer factoid natural language questions. Considering Wikipedia as a collection of about 5 million textual documents without relying on its graph structure, this setting brings the challenge of building systems able to do \textit{Machine Reading Comprehension} at scale. Most recent works (\cite{chen2017reading}, \cite{raison2018weaver}, \cite{min2018efficient}) explored the following pipeline to solve this task. First, retrieving dozens of documents using statistical methods (bigrams, tf-idf, BM25, etc.) or similarity search on dense representations between documents and questions \cite{karpukhin2020dense}, \cite{khattab2020colbert} and then applying a deep learning model trained for machine reading comprehension to find the answer. Some other works have been about designing methods to re-rank documents using more sophisticated methods like deep learning or reinforcement learning (\cite{wang2017r3}, \cite{lee2018ranking}). 

Yang et al. designed in BERTserini \cite{DBLP:journals/corr/abs-1902-01718} a pipeline of 2 steps: first, reducing the search space
thanks to the BM25 algorithm, extracting the text spans in each document retrieved with a finetuned BERT.
Eventually, the issue discussed in the introduction about scoring the relevance of spans of text coming from different paragraphs is tackled by taking the weighted average of the score from the BM25 algorithm and the score of the QA model. The weight is a hyperparameter tuned manually. 

Wang et al. \cite{wang2019multipassage} tackle the issue of comparing spans across several documents with a more sophisticated approach with \textit{Mutli-passage BERT}. They use a global normalization method: the retrieved passages are processed separately by the language model, and eventually, the $softmax$ operation is applied to all the tokens from all documents to normalize the scores and allow fair comparison between spans from several passages. They also use a passage re-ranker based on another BERT model: the BM25 algorithm retrieves 100 ranked passages,and the $softmax$ normalization is applied across the top 30 passages.

Seo et al. \cite{seo2018phraseindexed} introduced the \textit{Phrase-Indexed Question Answering} (PIQA) benchmark to make machine reading comprehension scalable. 

This benchmark enforces independent encoding of question and document answer candidates to reduce Question Answering to a simple similarity search task. Indeed, answer candidates are indexed offline. Closing the gap
between such systems and powerful models relying on query-aware context representation would be a great step toward solving the open-domain question-answering scalability challenge. The proposed baselines use
LSTM-encoders trained in an end-to-end fashion. While achieving encouraging results, the performances are far from state-of-the-art attention-based models.

\textit{DENSPI} \cite{seo-etal-2019-real} is the current state-of-the-art model on the PIQA benchmark. This system uses the BERT-large language model to train a siamese network to encode questions and index candidates independently. DENSPI is also evaluated on the squad-open benchmark. While being significantly faster than the other systems, it needs to be augmented by sparse representations of documents to be on par with them in terms of performance. 

\textit{Ocean-Q} \cite{fang2020accelerating}  proposes an interesting approach to solve the Open-Domain QA task by building an \textit{ocean} (a large set) of question-answer pairs using Question Generation and query-aware QA models. When asked, the most similar question from the \textit{ocean} is retrieved thanks to tokens similarity. This approach avoids the question-encoding step while being on par with previous models on the squad-open benchmark.

\section{Model}
In this section, the proposed model to solve the task is developed. 

\subsection{Pipeline Description}

The complete MIX pipeline is shown in figure \ref{pipeline}. It is made up of three fundamental building blocks. 

When a question is asked, we first select a few paragraphs relevant to the question (i.e., more likely to contain the correct answer). Later, we call this step the \textit{Retriever} module. It has to be highly efficient to tackle large corpora, with potentially several millions of paragraphs.

After that, we refine the retrieval step by re-ranking these paragraphs with a classification by relevance step, which we call the \textit{Scorer}. 

Finally, we extract from each paragraph the snippet that best answers the question. This is the \textit{Extractor} part.

\subsection{BM25 algorithm}

The \textit{Retriever} uses the BM25 algorithm, one of the most successful algorithms for textual information retrieval. It evaluates the relevance of each document relative to a query written in natural language. Indeed, when a request is made, the algorithm computes a score for each document in the dataset. This score is a sum of terms over the words in the question. Each term of the sum grows with the term frequency of the word in the document and is modulated by its inverse document frequency. paragraphs are finally sorted regarding their scores.

\subsection{Scoring paragraphs}

\begin{figure}[ht]
    \includegraphics[width = 0.45\textwidth]{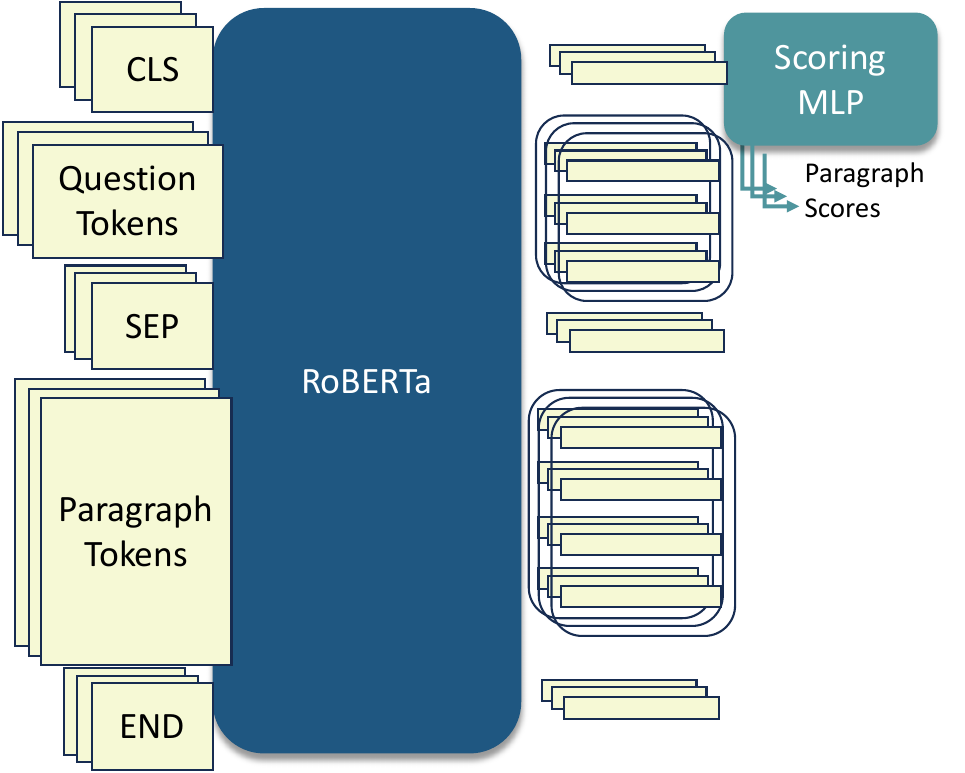}
\caption{The proposed scoring model. An example consists of a question and several candidate paragraphs. For each paragraph, the classification token embedding (CLS) is built to represent the given paragraph according to the question. Eventually relevance scores are computed from these vectors and normalized by the $softmax$ function.}
\label{scorer_model}
\end{figure}

The \textit{Scorer} allows to refine the classification of the paragraphs returned by the \textit{Retriever}. We use deep learning to implement this step, and we build a model that associates a relevance score to a pair (Question, Document) (figure \ref{scorer_model}). 
In this model, we use the classification token of the RoBERTa (base) \cite{liu2019roberta} language model to return the relevance score.

\subsection{Question Answering}

\begin{figure}[ht]
\centering
    \includegraphics[width = .45\textwidth]{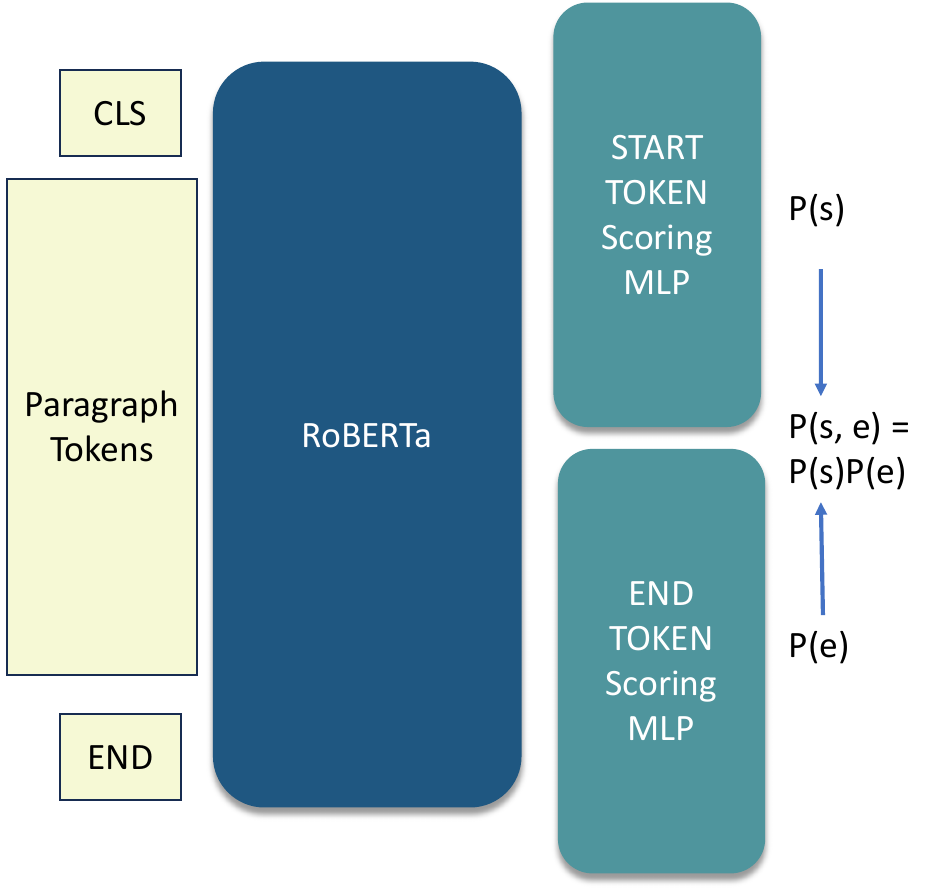}
\caption{The proposed QA model. The question and the paragraph are processed simultaneously by the language model to build question-aware paragraph representations, and then 2 dense layers allow us to find the beginning and the end of the answer.}
\label{question_answering_model}
\end{figure}

The \textit{Extractor} part of the pipeline uses a vanilla Question Answering model: the RoBERTa (base) language model finetuned to produce probability distributions on the paragraph tokens to identify the beginning and the end of the span of text answering an input question (figure \ref{question_answering_model}). 

\subsection{Multi-task Model}

As we have just seen, our system comprises two deep-learning models, one for the Scorer and the other for the Extractor.  These models solve the re-ranking of paragraphs and the QA tasks. This configuration can be heavy in terms of resources.

\begin{figure*}[ht]
    \includegraphics[width = \textwidth]{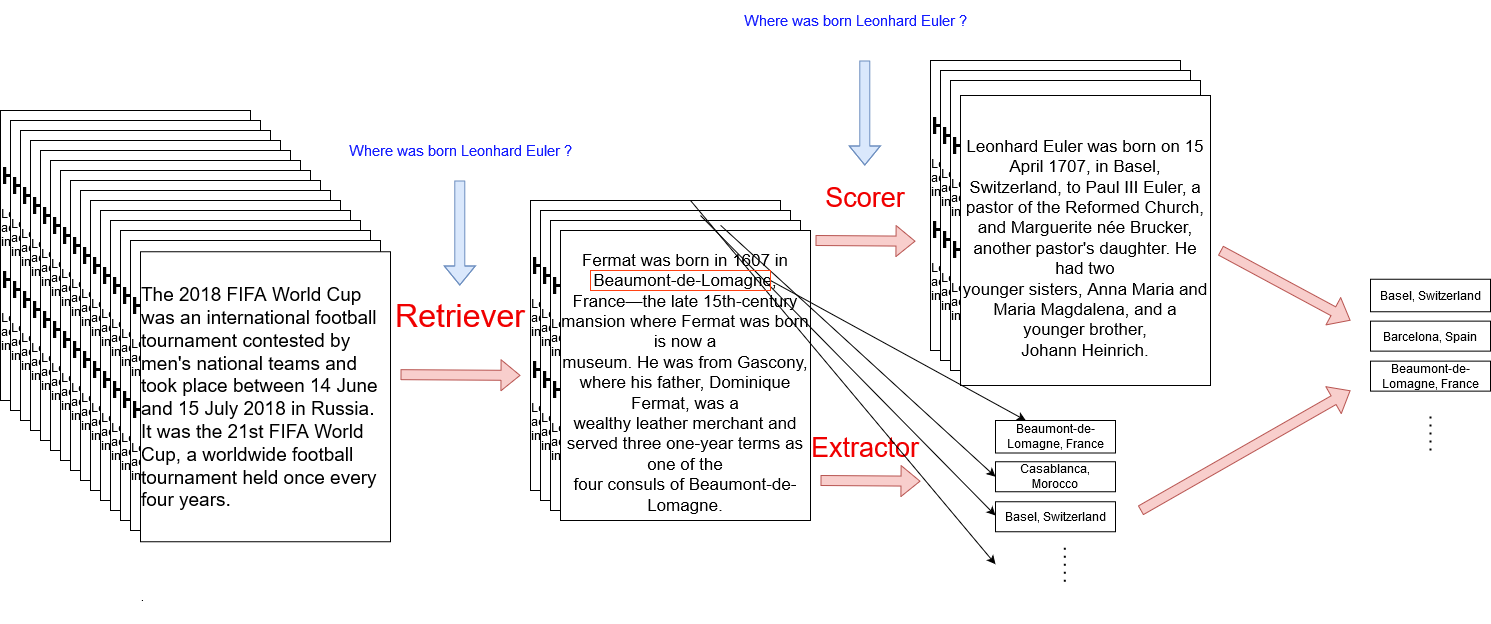}
\caption{The proposed multi-task pipeline. As before, the search space is first reduced thanks to the \textit{Retriever}, then the most relevant spans of text are extracted while the paragraphs are re-ranked. We can see that the question is processed only twice, once by the \textit{Retriever} and once by the multi-task model instead of three times in the previous pipeline.}
\label{pipeline2}
\end{figure*}

Since these two tasks are related in that they require understanding a text in the light of a question asked, a \textit{multitask learning} model could be designed instead. The goal is to learn both tasks by sharing part of their model parameters, allowing parallel classification of paragraphs and extraction of relevant spans of text (figure \ref{pipeline2}). This would save inference time and required memory. 

\begin{figure}[ht]
    \includegraphics[width = 0.48\textwidth]{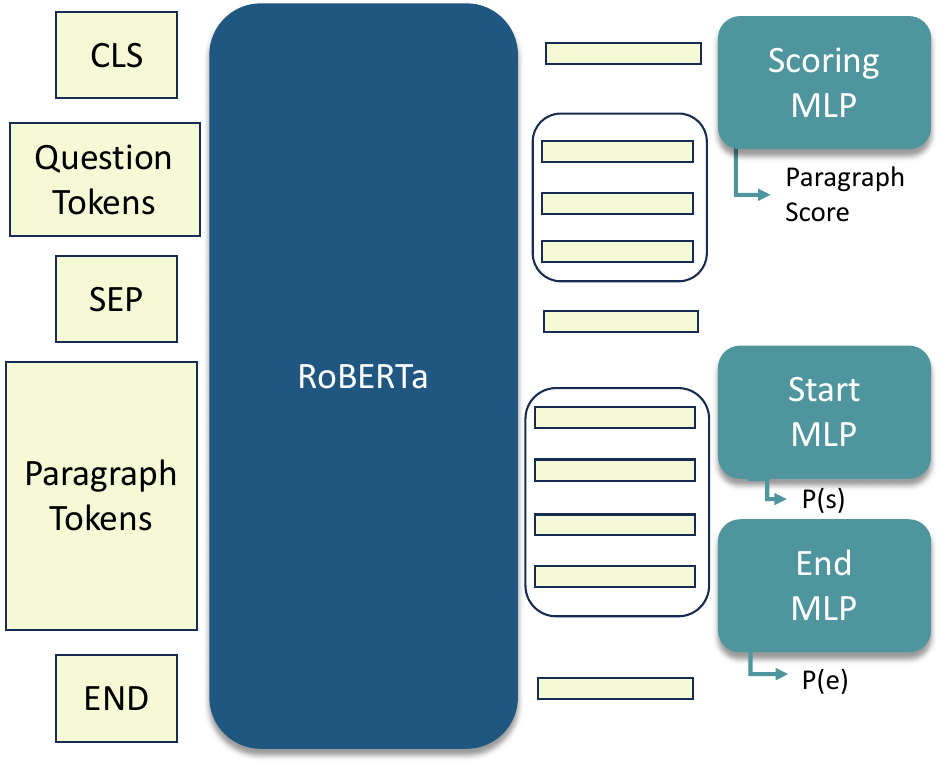}
\caption{The proposed Multi-task model. Now, the question-aware representations of the paragraph built by the language model allow us to perform both paragraph Scoring (through the CLS token) and Question Answering (through the paragraph tokens).}
\label{multitask_model}
\end{figure}

The proposed multi-task model is depicted in figure \ref{multitask_model}, where we can see that shared parameters are those of the language model (thus, the largest part of the set of parameters). We keep the layers specific to each task, i.e., the layer that takes the classification token as input in the \textit{Scorer} and the layers of start and end positions in the \textit{Extractor}. 

\paragraph{Search Engines} 
Our architecture is particularly adapted to a search engine setting where the user needs the results from different documents to be ranked by relevance according to a query. Indeed, our system can simultaneously extract the most relevant text spans in a batch of paragraphs while ranking the paragraphs themselves. Thus, the ranked paragraphs with their extracted answers can be displayed to the user efficiently.

\subsection{Training Objectives}

\subsubsection{Paragraphs Scoring}

When training our model to score paragraphs, we optimize the cross-entropy of the ground truth paragraph against the other ones. This corresponds to the following loss :

\begin{equation}
\begin{adjustbox}{max width=0.45\textwidth}
$
\mathcal{L}(\mathbf{\Theta}; Q, \mathcal{D}) = -score(d^*) + log\left(\sum_{d \in \mathcal{D}} exp\left(score(d)\right)\right)
$
\label{eq:scoring}
\end{adjustbox}
\end{equation}

with $\mathbf{\Theta}$ the parameters of the model, $Q$ the question, $\mathcal{D}$ the set of paragraphs returned by the Retriever and $d ^*$ the correct paragraph. 

\subsubsection{Question Answering}

Training the model for the Question Answering task is done by minimizing the cross entropy of the start and end positions of the correct answers.
Given a document $ D $ (\textit {tokenized} as $D = \{d_1, d_2, ..., d_n\}$), a question $Q$, the answer $A$ characterized by $ (s, e) $ its start and end positions in $D$, $\mathbf{\theta}$ the model parameters, $ \mathbf{P}_{start}$ the probability distribution for the start of the response and  $ \mathbf{P}_{end}$ the probability distribution for the end of the answer, we define the loss function $\mathcal{L}(Q, D; \mathbf{\theta})$ as the following (eq. \ref{eq:qa}):
\begin{equation}
\begin{adjustbox}{max width=0.45\textwidth}
$
\mathcal{L}(Q, D; \mathbf{\theta}) = -log\left(\mathbf{P}_{start}\left(s| Q, D; \theta\right)\right) -log\left(\mathbf{P}_{end}\left(e| Q, D; \theta\right)\right)
$
\label{eq:qa}
\end{adjustbox}
\end{equation}

\section{Experiment}
In this section, we show our experiments and results. 

\subsection{Data}

\paragraph{SQuAD v1.1 :}
SQuAD v1.1 \cite{rajpurkar2016squad} is a reading comprehension dataset comprising 100,000+ question-answers pairs from Wikipedia paragraphs. The span extraction part of our model is trained on the train set ($87599$ pairs) and evaluated on the development set ($10570$ pairs).                              
\paragraph{squad-open :}
\cite{chen2017reading} introduced the squad-open benchmark to tackle the Open-Domain question-answering setting. It uses the whole English Wikipedia as the unique source of knowledge and is evaluated on SQuAD v1.1 questions. The Wikipedia version is from 2016 and contains 5,075,182 articles.              
\subsection{Training and Implementation Details}

\subsubsection{Indexation}

We use the Python API of Elasticsearch \cite{10.5555/2904394} for document indexation. It allows us to take advantage of its native implementation of the BM25 algorithm. When indexing, we first used a sliding window with a stride of 300 tokens and considered paragraphs of 400 tokens. We end up indexing around 33,6 M paragraphs. \cite{wang2019multipassage} found that the value of granularity (i.e. the number of tokens considered to form a paragraph) impacts the performances of their system. Thus, we then test different granularity values and analyze their consequences below.  

We use Elasticsearch indexation to build the dataset to train the \textit{Scorer}: we apply the BM25 algorithm for each question in the dataset to all paragraphs in SQuAD. We only retrieve \textbf{30} paragraphs for each question. Each time the paragraph containing the ground truth answer is in the retrieved paragraphs, a new example is added to the dataset. Thus, an example in this set is represented by \textit{1 question} and a group of \textit{30 paragraphs}. We end up with 80k+ examples for the train set. To build the development set, we apply the same procedure to SQuAD-dev and end up with 10k+ examples in our development set.

\subsubsection{Training}

During the training, we alternate question-answering and Scoring optimization steps. We use \textit{AdamW} \cite{loshchilov2019decoupled} optimizer and learning rates of $5e-5$ and $1e-5$  with linear schedulers for Question Answering optimization and Scoring optimization, respectively. We set a batch size of 32 examples for Question Answering and a simulated batch size of 16 for scoring thanks to gradient accumulation. We also used Mixed Precision training  \cite{micikevicius2018mixed}, and the model was trained on a single 24GB GPU NVIDIA Quadro RTX 6000.

\subsection{Results}

\subsubsection{Metrics}

The performances of the QA model are evaluated thanks to classic exact-match (EM, exact overlap between ground truth span and prediction) and f1-score (F1, partial overlap between ground truth span and prediction). The performances of the Scorer are evaluated by observing if the paragraph containing the right answer is ranked first (precision @ 1).

\subsubsection{Results on SQuAD}

Table \ref{multitasksquad} provides results of our multitask model on the SQuAD dataset for both  Question Answering and re-ranking tasks. It shows that our model is on par with state-of-the-art performances in Question Answering.
In addition, the re-ranking task performs well as in 94.2 \% of the examples, the ground truth paragraph is ranked first.
The comparison with the finetuning of RoBERTa \cite{liu2019roberta} on the only task of question answering shows that the performances of our model do not suffer from multi-task learning.

\begin{table}[ht]
\begin{adjustbox}{max width=0.45\textwidth}
    \centering
    \begin{tabular}{|c|c|c|c|}
        \hline
         & EM & F1 & precision\\
         & & & @ 1\\
        \hline
        RoBERTa-base  & &  &  \\
        \cite{liu2019roberta} &  83.0 \%& 90.4\% & -\\
        on SQuAD (our run) & & &\\
        \hline
        multi-task & 83.5\% & 90.5\% & 94.2\% \\
        \hline
    \end{tabular}
    \end{adjustbox}
    \caption{Results of MIX on the QA and the scoring tasks. The vanilla Question Answering model is from a single run with a learning rate of 5e-5, linear scheduling, and a batch size of 16.}
    \label{multitasksquad}

\end{table}

\subsubsection{Results on the squad-open benchmark}

In this section, we present the results obtained on the benchmark squad-open. Table \ref{mix_perfs} provides the results of MIX on different numbers of paragraphs returned by the \textit{Retriever} (100 and 200) and for top 1, top 2, and top 3 snippets returned by the \textit{Extractor}. We observe that the model is more limited by the performance of the paragraph reclassification step than by the number of paragraphs returned by BM25. Indeed,  performances increase very quickly from top 1 to top 2 or from top 2 to top 3, while they increase less rapidly from 100 to 200 paragraphs.

\begin{table}[ht]
    \centering
    \begin{tabular}{|c|c|c|}
        \hline
          & exact-match & f1-score \\
        \hline 
        100 § / top 1 & 51.3 & 59.1\\
        \hline
        100 § / top 2 & 57.8 & 65.8\\
        \hline
        100 § / top 3 & \textbf{60.1} &\textbf{68.9}\\
        \hline
        200 § / top 1 & 52.8 & 60.9\\
        \hline
        
    \end{tabular}
    \caption{Results on the squad-open benchmark depending on the number of paragraphs retrieved by BM25 and the top k answers returned.}
    \label{mix_perfs}
\end{table}

\paragraph{Different levels of granularity}

\begin{figure}[ht]
\centering
    \includegraphics[width = .45\textwidth]{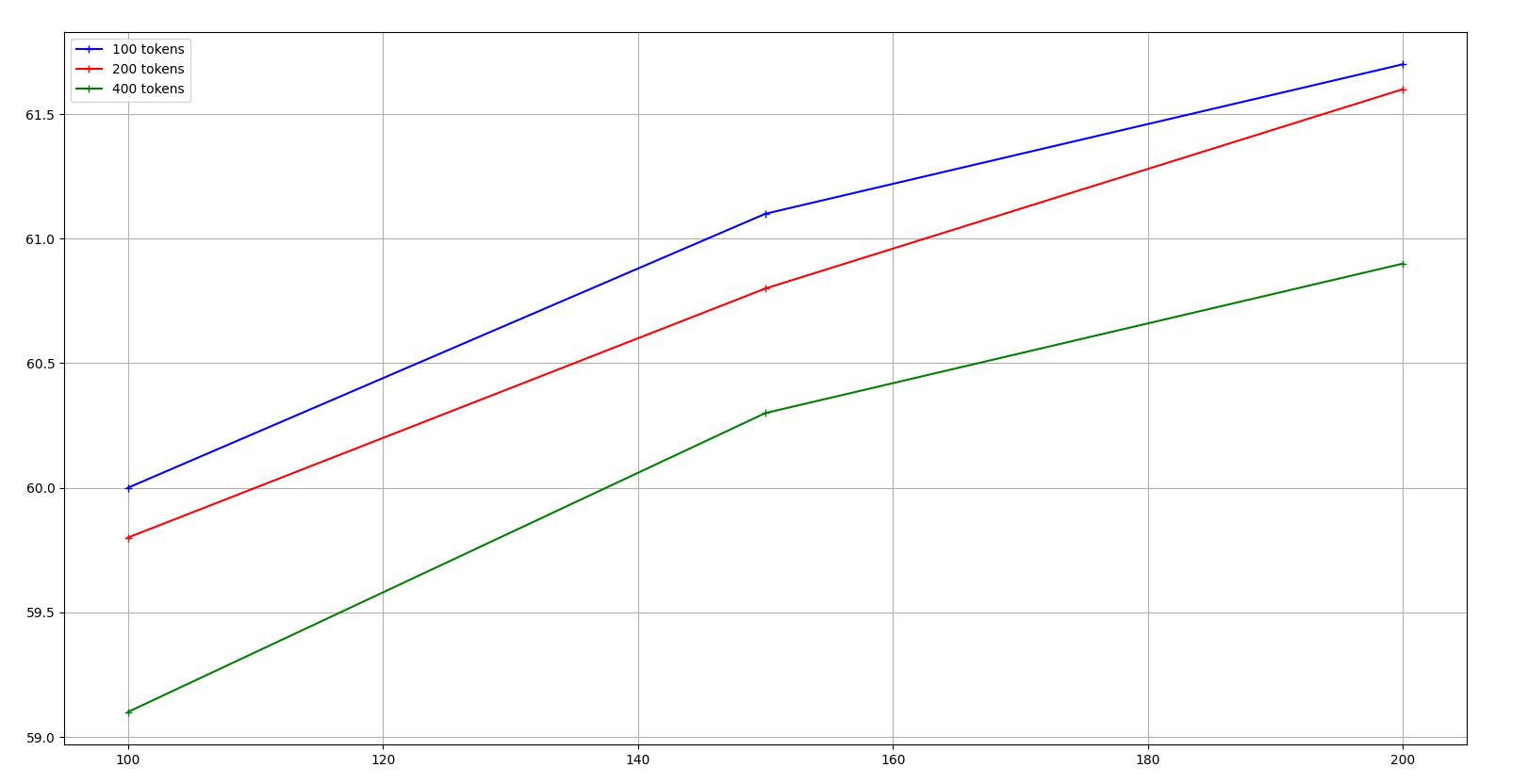}
\caption{f1-score of the system on the squad-open benchmark as a function of granularity and number of paragraphs retrieved (100, 150, and 200). the green curve represents the performances of the 400 tokens level (stride of 300), the red curve represents the 200 hundred tokens level (stride of 100) and the blue curve the 100 tokens level (stride of 50). We see that reducing the number of tokens per paragraph significantly improves the performance of the system.}
\label{granularity}
\end{figure}

\cite{wang2019multipassage} explored numerous levels of granularity when indexing paragraphs. They found the counterintuitive result that passages of 100 tokens provided optimal performances. Thus, we also test different granularity levels: 100, 200, and 400. Figure \ref{granularity} shows that the granularity level of 100 tokens is constantly better than the other ones so it confirms \cite{wang2019multipassage} results.

\paragraph{Comparison with state-of-the-art}

We also compare the performance of MIX to other state-of-the-art models evaluated on the squad-open benchmark (table \ref{squad-open-results}). Our system obtains state-of-the-art performances, beating \textit{Multi-Passage BERT} by 2.5 points and 2.7 points in exact-math and f1-score respectively, for their BERT-base based system and by 0.8 points in both metrics for their BERT-large based system while our system is based on RoBERTa-base. These results show that our system while being simple conceptually and using only one language model, can overcome the issue of comparing spans of texts across different paragraphs.

\begin{table}[ht]
\begin{adjustbox}{max width=0.48\textwidth}
    \centering
\begin{tabular}{cccc}
    \hline
     & EM & F1 \\
    \hline
     DrQA \small\cite{chen2017reading} & 29.8 & - \\
    
    R$^3$ \small\cite{wang2017r3} & 29.1 & 37.5 \\
    
    Paragraph ranker \small\cite{lee2018ranking} & 30.2 & - \\

    MINIMAL \small\cite{min2018efficient} & 34.7 & 42.5 \\
    
    Weaver \small\cite{raison2018weaver} & - & 42.3 \\
    
    Multi-step reasoner \small\cite{das2018multistep} & 31.9 & 39.2 \\
    
    BERTSERINI \small\cite{DBLP:journals/corr/abs-1902-01718} & 38.6 & 46.1 \\
    
    \underline{Multi-passage BERT (base)} \small\cite{wang2019multipassage} & \underline{51.2} & \underline{59.0}\\
    
    \underline{Multi-passage BERT (large)} \small\cite{wang2019multipassage} & \underline{53.0} & \underline{60.9}\\
    
    DENSPI (Hybrid) \small{\cite{seo-etal-2019-real} (PIQA method)} & 36.2 & 44.4 \\
    
    DENSPI (Dense Only) \small{\cite{seo-etal-2019-real} (PIQA method)} & 20.5 & 13.3  \\

    Ocean-Q \small{\cite{fang2020accelerating} (PIQA method)} & 32.7 & 39.4 &\\

    \textbf{MIX} \small{(ours) (200 paragraphs of length 100)} & \textbf{53.8} & \textbf{61.7} \\

    \hline
\end{tabular}
\end{adjustbox}
    \caption{Results on the squad-open benchmark.}
    \label{squad-open-results}
\end{table}

\section{Conclusion}

MIX, a multi-task learning approach to solve open-domain question answering, relying on the  BM25  algorithm as a \textit{Retriever} to reduce the search space and the powerful RoBERTa language model finetuned to achieve both paragraph re-ranking (\textit{Scorer}) and spans of text extraction (\textit{Extractor}). Our system achieves state-of-the-art results on the squad-open benchmark. Thus, we showed that a simple pipeline consisting of a retriever, a re-ranker, and a reader is sufficient to overcome the issue of comparing 2 possible answers from different paragraphs. The novelty of our paper comes from using multi-task learning to allow us to perform simultaneous re-ranking and reading comprehension. Our system is particularly adapted to a search engine setting where several results ranked by relevance are provided to the user. Our evaluation shows that the results are more limited by the performance of the \textit{Scorer} than by the \textit{Retriever} that reduces the search space of millions of paragraphs to a hundred, so future research might be directed to the improvement of the \textit{Scorer} performances. Furthermore, we confirm the finding of \cite{wang2019multipassage}, according to which shorter paragraphs allow better performances on the open-domain QA task. Eventually, further improvements might be achieved using a heavier language model such as RoBERTa-large.

\bibliography{main.bib}
\bibliographystyle{acl_natbib}

\end{document}